\documentclass[runningheads]{llncs}
\usepackage[T1]{fontenc}
\usepackage[breaklinks,colorlinks]{hyperref} 
\usepackage{amssymb, amsmath}
\usepackage{subcaption} 
\usepackage{graphicx}
\usepackage{booktabs}
\usepackage{orcidlink}
\usepackage{multirow}
\usepackage{ulem}
\usepackage{cleveref}
\usepackage{cite}

\usepackage{color}
\definecolor{darkgreen}{RGB}{20,158,30}
\definecolor{darkred}{RGB}{223,35,20}
\definecolor{darkblue}{RGB}{23,35,220}
\begin{document}
\title{RADA: Robust and Accurate Feature Learning with Domain Adaptation}
\titlerunning{RADA}

\author{Jingtai He \and
Gehao Zhang\and
Tingting Liu \and
Songlin Du}
\authorrunning{J. He et al.}
\institute{School of Automation, Southeast University, Nanjing 210096, China\\
\email{\{hejingtai,sdu\}@seu.edu.cn}}
%
\maketitle              
\begin{abstract}
Recent advancements in keypoint detection and descriptor extraction have shown impressive performance in local feature learning tasks. However, existing methods generally exhibit suboptimal performance under extreme conditions such as significant appearance changes and domain shifts. In this study, we introduce a multi-level feature aggregation network that incorporates two pivotal components to facilitate the learning of robust and accurate features with domain adaptation. First, we employ domain adaptation supervision to align high-level feature distributions across different domains to achieve invariant domain representations. Second, we propose a Transformer-based booster that enhances descriptor robustness by integrating visual and geometric information through wave position encoding concepts, effectively handling complex conditions. To ensure the accuracy and robustness of features, we adopt a hierarchical architecture to capture comprehensive information and apply meticulous targeted supervision to keypoint detection, descriptor extraction, and their coupled processing. Extensive experiments demonstrate that our method, RADA, achieves excellent results in image matching, camera pose estimation, and visual localization tasks.
\end{abstract}
%
%
\section{Introduction}
Robust and accurate keypoint detection and descriptor extraction are fundamental for various computer vision tasks, such as visual localization~\cite{toft2020long}, and structure from motion (SfM)~\cite{RCM_ECCV24}. With the advent of deep learning, significant advancements~\cite{detone2018superpoint,dusmanu2019d2,luo2020aslfeat} have been achieved, enhancing both the accuracy and efficiency of local features compared to traditional hand-crafted methods~\cite{lowe2004distinctive,rublee2011orb}.

The adoption of sparse-to-sparse feature learning, which follows a `detect-then-describe' paradigm, has proven common and effective in feature matching methodologies. Numerous methods adhering to this paradigm have been developed, focusing on optimizing individual components, such as keypoint detection or descriptor extraction, to enhance the local features. Nevertheless, the robustness of these features can significantly deteriorate under extreme domain variations, such as day-night changes or different seasons. This deterioration occurs due to the susceptibility of the keypoint detection stage to changes in low-level image statistics, affecting the subsequent feature descriptors' ability to maintain invariance. To mitigate this challenge, we draw inspiration from transfer learning~\cite{ganin2015unsupervised} and propose domain adaptation supervision implemented prior to the keypoint detection stage. This component aims to establish a mapping between different domains, thereby minimizing discrepancies in high-level feature distributions. By aligning these distributions, we improve keypoint detector stability, the robustness and invarance of descriptors across varying conditions.

Under large illumination and viewpoint changes, local visual information becomes unreliable and indistinguishable, leading to the degradation of descriptors. To address this issue, the global modeling capability of Transformer has been increasingly utilized to incorporate contextual information into feature descriptors, enhancing their robustness~\cite{luo2019contextdesc,wang2022mtldesc,wang2023featurebooster}. 
Motivated by these advancements, we develop a Transformer-based booster designed to boost the robustness and accuracy of descriptors by integrating both the visual and geometric information from the whole local features. A noteworthy aspect of our approach is the use of a wave-based position encoder, which offers enhanced encoding capabilities. Leveraging the expansive receptive field inherent to the Transformer architecture, the boosted descriptors become more robust and discriminative by encapsulating global spatial contextual information.

\begin{figure}[t]
	\centering
	\includegraphics[width=0.75\linewidth]{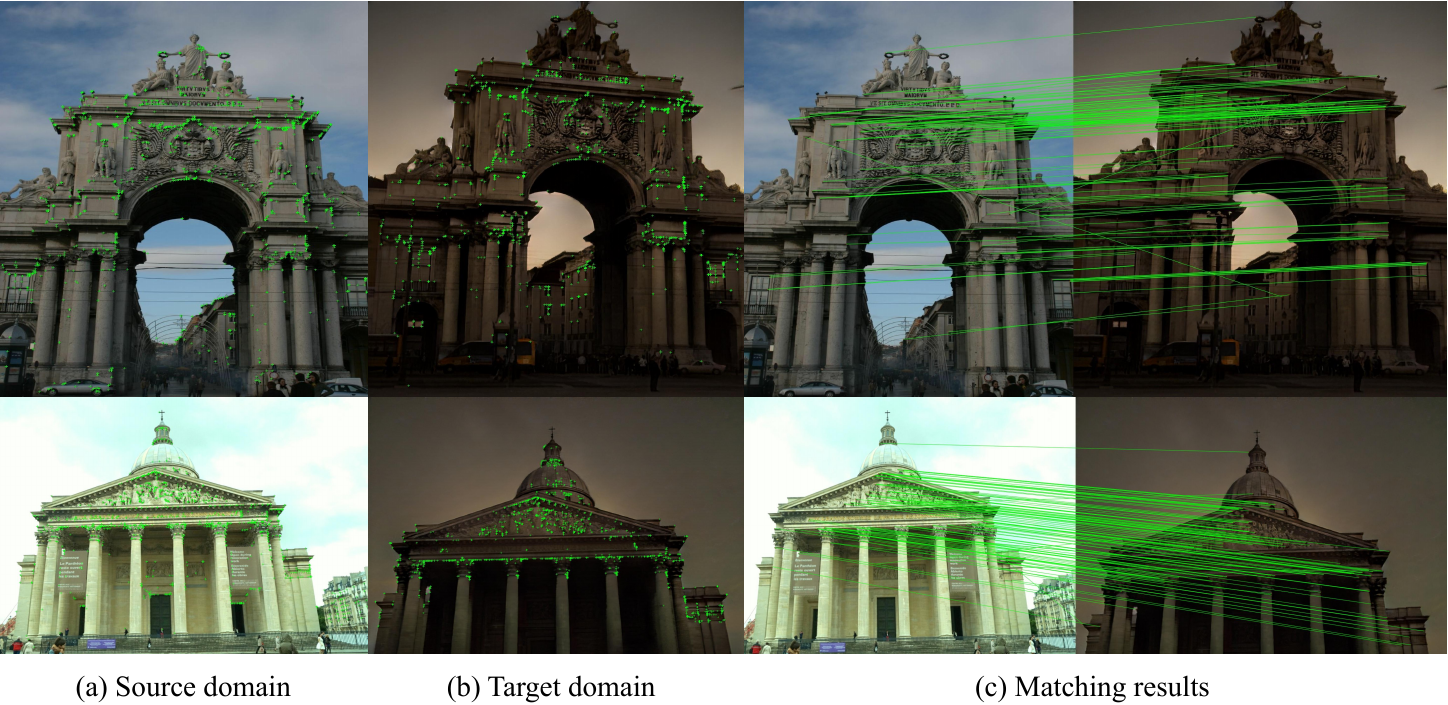}
	\caption{
	 Visualization of the detected keypoints and the matches using our RADA on the Megadepth~\cite{li2018megadepth} validation set.  
	(a) Keypoints detection from the source domain (daytime). 
	(b) Keypoints detection from the target domain (nighttime).
        (c) The matching results between image pairs from different source and target domains. The green lines show correct correspondences.}
	\label{matching}
\end{figure}

In summary, the contributions of our paper include:
\begin{itemize}     
  \item [-]
  We introduce domain adaptation supervision to align high-level feature distributions across different domains and achieve invariant representations using transfer learning concepts. 
  \item [-]
  We propose a Transformer-based booster to enhance descriptor robustness by integrating the global descriptors and positional information through the concepts of wave amplitude and phase.
  \item [-] 
  We develop a hierarchical feature aggregation network incorporating these two key components with carefully designed loss functions to achieve \textbf{R}obust and \textbf{A}ccurate feature learning with \textbf{D}omain \textbf{A}daptation (RADA).
\end{itemize}

%
%
\section{Related Works}

\subsection{Local Feature Learning}
Descriptors were predominantly hand-crafted initially, with SIFT and ORB~\cite{lowe2004distinctive} being the most representative examples. Then deep learning has significantly advanced the performance of learned local descriptors. Most patch-based learning descriptors~\cite{tian2019sosnet} adopt the network architecture proposed in L2-Net~\cite{Tian_2017_CVPR} with various loss functions. However, these methods primarily concentrate on descriptor extraction, with the receptive fields confined to the image patches. Other descriptors~\cite{zhao2022alike}, following the `detect-then-describe' paradigm, estimate score maps and descriptor maps, where the score map indicates keypoint probability. ALIKE~\cite{zhao2022alike} introduces a differentiable keypoint detection method based on the score map, enabling backpropagation of gradients and subpixel-level keypoint production. Additionally, joint detection and description methods have shown increasing performance. SuperPoint~\cite{detone2018superpoint} proposes a network trained on homography image pairs. R2D2~\cite{revaud2019r2d2} integrates grid peak detection with reliability prediction for repeatable keypoints. DISK~\cite{tyszkiewicz2020disk} trains score and descriptor maps via reinforcement learning. D2-Net~\cite{dusmanu2019d2} detects keypoints using channel and spatial maxima on low-resolution feature maps. ASLFeat~\cite{luo2020aslfeat} uses hierarchical features to detect keypoints and capture invariant descriptors with deformable convolutions.

Despite numerous related works focusing on learning local features, there is limited attention on enhancing descriptor robustness and accuracy through domain adaptation and the corresponding implementation techniques.

\subsection{Domain Adaptation}
Domain adaptation endeavors to mitigate the disparity between features learned from complex networks across source and target data. This concept has been extensively developed through deep learning, enhancing adaptation performance in computer vision and multimedia applications~\cite{tzeng2017adversarial}. Several methods~\cite{tzeng2014deep} employ the Maximum Mean Discrepancy (MMD)~\cite{quinonero2008covariate} loss, which computes the norm of the difference between two domain means, to minimize the difference of their feature distributions. Other approaches~\cite{tzeng2017adversarial} motivate from adversarial learning, with the gradient reversal algorithm (ReverseGrad) from DANN~\cite{ganin2015unsupervised} being particularly relevant.This algorithm approaches domain invariance as a binary classification problem, confusing the domain classifier by inverting its gradients and maximizing its loss to learn an invariant representation that is both discriminative and indistinguishable across different domains.

In local feature learning tasks, DomainFeat~\cite{xu2023domainfeat} is the first to introduce domain adaptation to the field, developing image-level domain invariance supervision by fusing domain-invariant representations. Motivated by these advancements, we implement our domain adaptation supervision, which organically combines the MMD loss and gradient reversal algorithm to learn accurate and robust features across different domains.

\subsection{Feature Context Awareness}
The spatial configuration of keypoints and descriptors of an image defines a global feature context, and incorporating this contextual information into descriptors has become a growing trend in local feature matching methods~\cite{sarlin2020superglue}. ContextDesc~\cite{luo2019contextdesc} uses keypoint positions, original and high-level local features to encode geometric and visual contexts. MTLDesc~\cite{wang2022mtldesc} incorporates non-local contextual information to capture high-level dependencies among distant features by employing an adaptive global context enhancement module in conjunction with multiple local enhancement modules. FeatureBooster~\cite{wang2023featurebooster} takes only the descriptors and keypoint positions within images as inputs and employs a lightweight Transformer to aggregate this information into new boosted descriptors.

However, most of these works use MLPs as position encoders, which have relatively weak encoding abilities. Inspired by Wave-MLP~\cite{tang2022image,lu2023paraformer}, we recognize that phase information is equally important as amplitude information in vectors. In the Transformer-based booster, we use the Euler formula to combine the amplitude information of the descriptor with the phase information of the position. This fusion process produces position-aware descriptors.

%
%
\section{Methodology}

\begin{figure}[t]
	\centering
	\includegraphics[width=11.9cm]{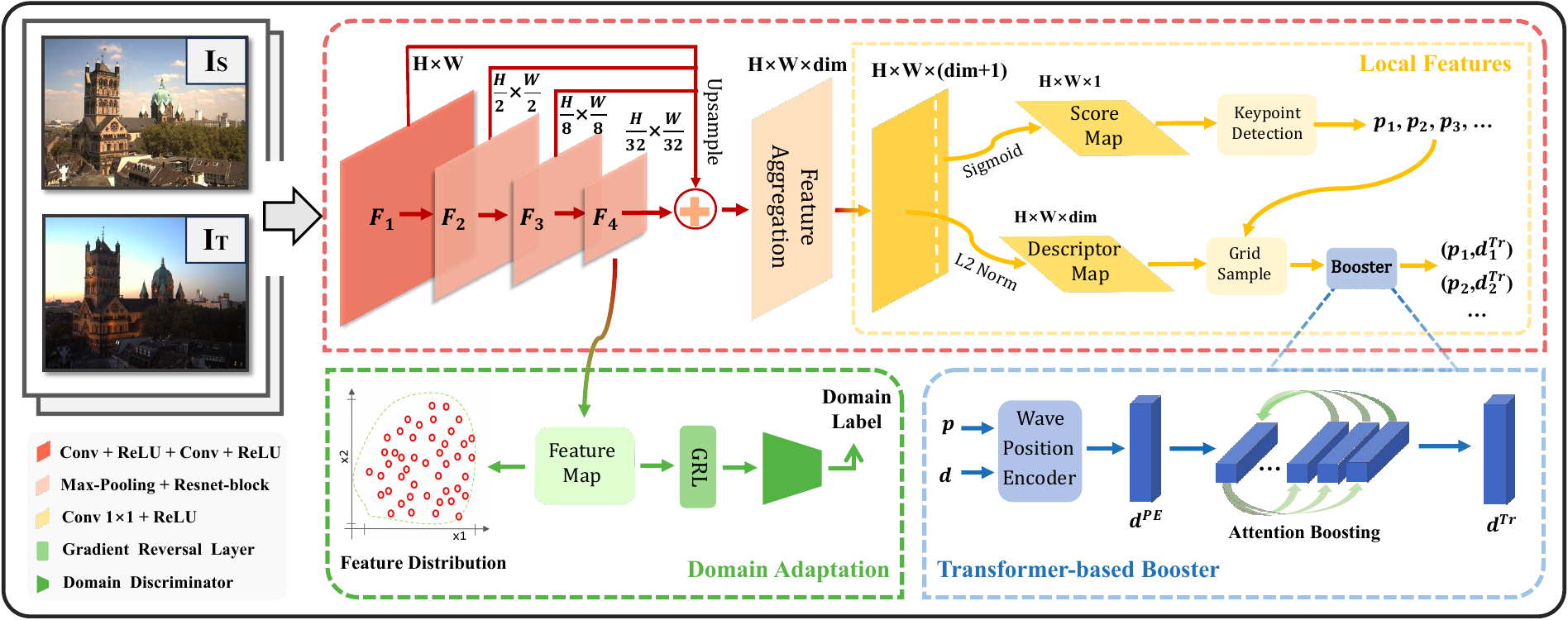}
	\caption{
	 Network architecture. Our RADA consists of three components: keypoint detection and descriptor extraction backbone to learn accurate local features from hierarchical feature maps, domain adaptation supervision to achieve domain-invariant representations, and a Transformer-based booster to improve the robustness of descriptors. The cross-domain training image pairs with ground truth are produced from Megadepth~\cite{li2018megadepth}.}
	\label{network}
\end{figure}

\subsection{Backbone}
\label{sec:backbone}

The procedure of our network to learn global domain-invariant features includes feature encoding, feature aggregation, and local feature extraction. Starting from an image $\textbf{\textit{I}} \in \mathbb{R}^{H \times W \times 3}$, the backbone process proceeds to generate a score map $\textbf{\textit{S}} \in \mathbb{R}^{H \times W}$ and a dense descriptor map $\textbf{\textit{D}} \in \mathbb{R}^{H \times W \times dim}$, and finally identifies the keypoints $\textbf{\textit{p}} \in \mathbb{R}^{2}$ along with their corresponding descriptors $\textbf{\textit{d}} \in \mathbb{R}^{dim}$.

\noindent
\textbf{Feature Encoding:} The feature encoder converts the image \textbf{\textit{I}} into multi-scale features \textbf{\textit{F$_1$}}, \textbf{\textit{F$_2$}} ,\textbf{\textit{F$_3$}}, and \textbf{\textit{F$_4$}} employing four neural network blocks with channel sizes of 32, 64, 128, and 128, respectively. The first block extracts low-level image features through a two-layer 3×3 convolutional operation followed by a ReLU activation function. The subsequent three blocks consist of a max-pooling layer followed by a 3×3 ResNet operation, with the downsampling rate of max-pooling set to 0.5 in block 2 and further reduced to 0.25 in blocks 3 and 4.

\noindent
\textbf{Feature Aggregation:} This component aims to combine multi-scale features \textbf{\textit{F$_1$}}, \textbf{\textit{F$_2$}} ,\textbf{\textit{F$_3$}}, and \textbf{\textit{F$_4$}} to enhance both localization and representation capabilities. We use a 1×1 convolution and a bilinear upsampling layer for the hierarchical features to align their dimensions and resolutions. By concatenating these aligned features, we acquire the final aggregated feature \textbf{\textit{F}} for the subsequent stages of keypoint detection and descriptor extraction.

\noindent
\textbf{Feature Extraction:} The final local feature learner first produces an $H\times W\times (dim+1)$ feature map consisting of two main components through a 1x1 convolution layer with ReLU activation. Then the descriptor map \textbf{\textit{D}} is generated by applying L2 normalization to the first $dim$ channels, while the score map \textbf{\textit{S}} is obtained by applying Sigmoid activation to the last channel.
A Differentiable Keypoint Detection (DKD) module~\cite{zhao2022alike} is applied to obtain trainable and differentiable keypoints by recognizing the local maxima utilizing non-maximum suppression (NMS) on the score map \textbf{\textit{S}}. Then positions of the keypoints are refined using softargmax function on the local patches to extract differentiable subpixel keypoints.
Based on the accurate keypoints, the network samples their corresponding descriptors in a grid-like manner from the descriptor map \textbf{\textit{D}}. Finally, the keypoints and descriptors are further enhanced through the Transformer-based booster described in~\cref{sec:boosting}, resulting in accurate local features.

\subsection{Domain Adaptation Supervision}
\label{sec:DA}

For the image \textit{\textbf{I$_S$}} from source domain and the image \textit{\textbf{I$_T$}} from target domain, their inconsistency between the feature domains can disrupt the processing of keypoint detection and descriptor extraction. 

\begin{figure}[tb]
  \centering
  \begin{minipage}{0.92\textwidth}
  \begin{subfigure}{0.54\linewidth}
    \includegraphics[width=\linewidth]{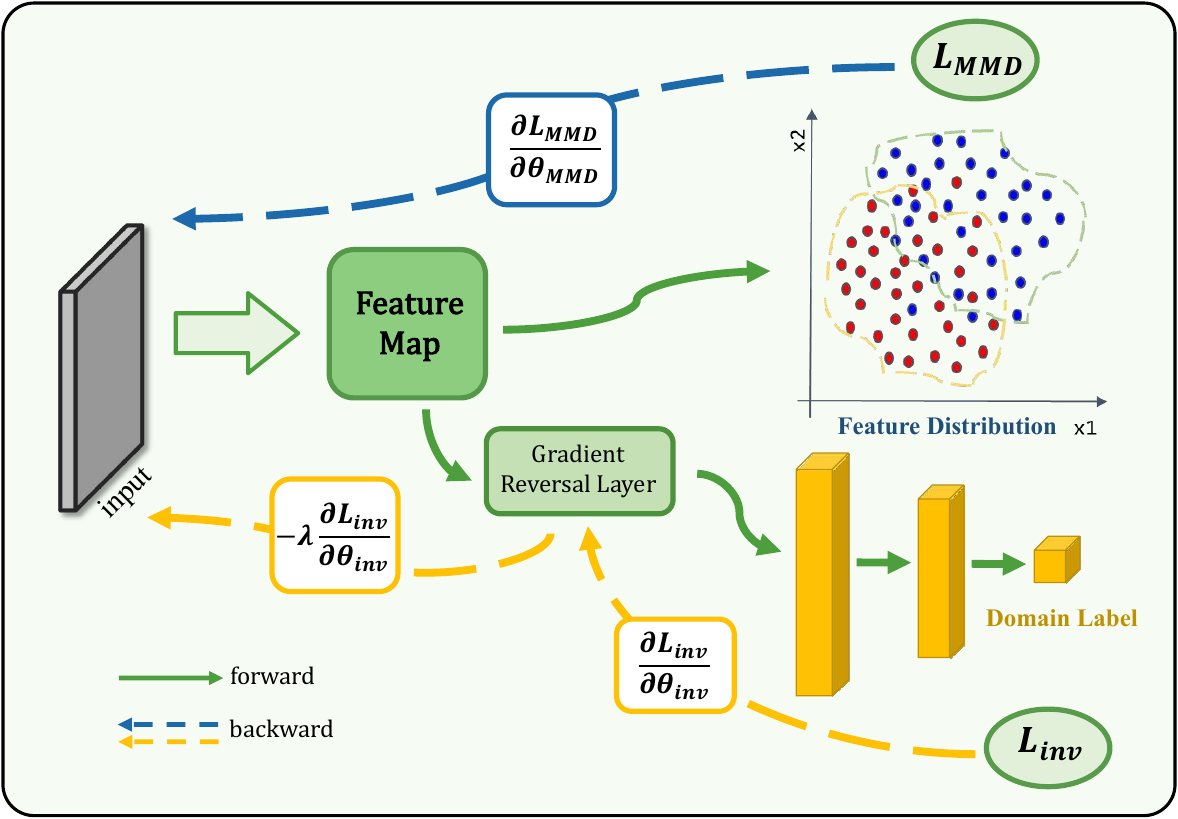}
    \caption{Domain Adaptation Supervision}
    \label{fig:DA}
  \end{subfigure}
  \hfill
  \begin{subfigure}{0.34\linewidth}
    \includegraphics[width=\linewidth]{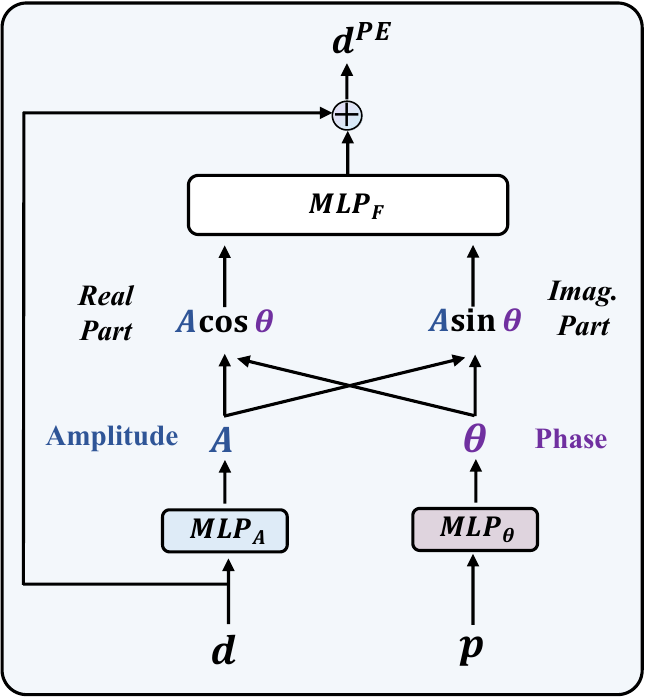}
    \caption{Wave Position Encoder}
    \label{fig:WavePE}
  \end{subfigure}
  \end{minipage}
  \caption{
    Clear illustration of the two modules. (a) Domain adaptation is achieved by introducing two branched tasks after the feature map. (b) Wave-PE combines the amplitude and the phase encoded by the information of local features.}
\end{figure}

As shown in Fig.~\ref{fig:DA}, to align the high-level feature distributions between the paired images, we first utilize the Maximum Mean Discrepancy (MMD) metric~\cite{quinonero2008covariate} with a linear kernel function. This distance is computed based on a particular representation, \textbf{$\phi$}, operating on source domain high-level features, $x_s \in X_S$, and target domain high-level features, $x_t \in X_T$. The empirical approximation of this distance is computed as follows:
\begin{equation}
  \mathcal{L}_{\mathrm{MMD}}(X_S, X_T) =  \left\| \frac{1}{|X_S|} \sum_{x_s \in X_S} \phi(x_s) - \frac{1}{|X_T|} \sum_{x_t \in X_T} \phi(x_t) \right\|.
\end{equation}

Next, to achieve the domain-invariant representations, we apply a gradient reversal layer (GRL)~\cite{ganin2015unsupervised} after the high-level feature map, which makes the adversarial discriminator unable to distinguish between different domains. This is followed by fully connected layers with channels to ($\text{dim}$, 512, 128), culminating in a binary classifier. The domain classifier identifies the invariant domain representation features $(X_S, X_T)$ through minimizing the cross-entropy loss:
\begin{equation}
  \mathcal{L}_{\mathrm{adv}}(X_S, X_T) = \frac{1}{N} \sum_{i=1}^{N} \left( -l_i \log(s_i) - (1 - l_i) \log(1 - s_i) \right),
\end{equation}
where $l_i$ is a boolean indicating the domain classification label: $l_i = 1$ for the target domain and $l_i = 0$ for the source domain. The domain membership prediction score for the target image is represented by $s_i \in \mathbb{R}$.

During forward propagation, the gradient of the features through the GRL remains unchanged. Conversely, during backpropagation, the GRL reverses the gradient, thereby maximizing the loss and updating the parameters in the negative direction of the gradient. It minimizes the distribution distance and maximizes the domain classifier loss, leading to the extraction of domain-invariant features. This process weakens the discrimination between the source and target images for the discriminator network, mapping the feature domains of the two images as closely as possible, thereby improving network performance.
Finally, we define the loss function of the domain adaptation supervision as:
\begin{equation}
\label{eq:da}
 \mathcal{L}_{\mathrm{da}} = \mathcal{L}_{\mathrm{adv}} + \lambda \mathcal{L}_{\mathrm{MMD}}.
\end{equation}

\subsection{Transformer-based Booster}
\label{sec:boosting}

Given $N$ keypoints are detected in the image \textbf{\textit{I}} through the network, their positions \textbf{\textit{p $\in \mathbb{R}^{N \times 3}$}} and descriptors \textbf{\textit{d}} $\in \mathbb{R}^{N \times C}$ are dynamically fused using a Wave Position Encoder (Wave-PE) to enhance descriptors robustness. The global context, encompassing description of other features along with the spatial arrangement of all local features, is then integrated by a Transformer. This incorporation of global information significantly enhances the robustness of individual descriptors, resulting in improved performance under challenging domain variations.

\subsubsection{Wave Position Encoder} strengthens descriptors with more information by fusing visual and geometric information through amplitude and phase relationships. Compared with traditional MLP-based position encoders which aggregate feature descriptors with fixed weights (positions), this method enhances encoding capacity by dynamically fusing feature representations through wave operations, such as decomposition and superposition. Specifically, the amplitude is the real value representing each extracted descriptor, while the phase term, a unit complex value, modulates the relationship between descriptors and positions within the MLP. The phase difference between these wave-like descriptors affects their aggregated output, with descriptors having close phases tending to enhance each other in subsequent attention calculations due to higher similarity.

In Wave-PE, the position encoding is achieved through the application of the Euler formula, which decomposes wave components $\tilde{\textit{\textbf{w}}}$ characterized by amplitude \textbf{\textit{A}} and phase $\boldsymbol{\theta}$ into their respective real and imaginary constituents, thereby facilitating efficient processing:
\begin{equation}
    \tilde{w}_j = A_j \odot e^{i \theta_j} = A_j \odot \cos \theta_j + i \cdot A_j \odot \sin \theta_j, \quad j = 1, 2, \ldots, N.
\end{equation}

As illustrated in Fig.~\ref{fig:WavePE}, the amplitude and phase information are encoded by two learnable networks based on the extracted descriptors and positions. Another learnable network fuses the complex representations into position encoding:
\begin{equation}
\label{eq:wavePE}
\begin{aligned}
    & A_j = \mathrm{MLP}_A(d_j), \quad \theta_j = \mathrm{MLP}_{\theta}(p_j), \\
    & d^{\mathrm{PE}}_j = d_j + \mathrm{MLP}_F \left( [A_j \odot \cos \theta_j, A_j \odot \sin \theta_j] \right),
\end{aligned}
\end{equation}
where $[\cdot, \cdot]$ represents concatenation. For simplicity, a two-layer MLP is used for the learnable networks in~\cref{eq:wavePE}.

\subsubsection{Attention-Free Transformer} (AFT) offers lower time and space complexity compared to the traditional multi-head attention (MHA) mechanism used in vanilla Transformers. Inspired by FeatureBooster~\cite{wang2023featurebooster}, we employ the AFT-attention~\cite{zhai2021attention} to aggregate all local feature information, thereby forming a global context that enhances descriptor robustness. AFT rearranges the computational sequence of the basic three quantities \textbf{\textit{Q}}, \textbf{\textit{K}}, \textbf{\textit{V}}, performing element-wise multiplication of \textbf{\textit{k}} and \textbf{\textit{v}} rather than traditional matrix dot products. The Attention-Free Transformer performs the following operation for the given keypoint \textbf{\textit{X}}$_i$:
\begin{equation}
\begin{aligned}
    f_i(X) &= \sigma(Q_i) \odot \frac{\sum_{j=1}^N \exp(K_j) \odot V_j}{\sum_{j=1}^N \exp(K_j)} \\
    &= \sigma(Q_i) \odot \sum_{j=1}^N (\mathrm{softmax}(K) \odot V)_j,
\end{aligned}
\end{equation}
where $\sigma(\cdot)$ is the nonlinearity applied to the query using Sigmoid function; The subscripts $i$ and $j$ index the matrix row of \textbf{\textit{Q}}, \textbf{\textit{K}}, \textbf{\textit{V}} respectively. AFT implements a modified version of MHA, in which the attention heads are as many as feature dimensions. In this approach, the attention score in MHA is replaced by element-wise multiplication and subsequent normalization of \textbf{\textit{k}} and \textbf{\textit{v}}, simulating their contribution to the current \textbf{\textit{q}}. This allows attention to be evaluated by preserving the global interactions between queries and values, without the need to compute the full attention matrix. As a result, the time and space complexity is comparable to that of linearized attention with respect to feature size. Due to the strong global modeling capacity and less computation overhead, the distinguishability of local features can be enhanced effectively.

Following previous work~\cite{revaud2019r2d2,wang2023featurebooster}, we approach the feature matching as a nearest neighbor retrieval issue and employ Average Precision (AP) for evaluation~\cite{boyd2013area}. The attention-boosted local feature descriptors, following the processing, $\{$\textbf{\textit{d}}$^{PE}_1$,\textbf{\textit{d}}$^{PE}_2$, $\ldots$ ,\textbf{\textit{d}}$^{PE}_N \}$ $\xrightarrow{}$ \textbf{\textit{d}}$^{Tr}_j$, are optimized to maximize AP for all descriptors \textbf{\textit{d}$^{Tr}$}, with the training goal to minimize the following loss function:
\begin{equation}
    \mathcal{L}_{\mathrm{Tr}} = 1 - \frac{1}{N} \sum_{i=1}^N \mathrm{AP}(d^{\mathrm{Tr}}_j).
\end{equation}

\subsection{The Loss Functions}
In addition to the domain adaptation loss and Transformer-based boosting loss previously mentioned, we define the detector loss, descriptor loss and coupling loss for the overall supervision during training.

Given an image pair \textit{\textbf{I}}$_A$ and \textit{\textbf{I}}$_B$, our hierarchical network learns the score maps and descriptor maps. The keypoints \textbf{\textit{P}}$_A \in \mathbb{R}^{N_A \times 2}$ and \textbf{\textit{P}}$_B \in \mathbb{R}^{N_B \times 2}$ are then detected by the DKD module from their respective score maps. The corresponding descriptors for these keypoints are extracted from their sets \textbf{\textit{D}}$_A \in \mathbb{R}^{N_A \times dim}$ and \textbf{\textit{D}}$_B \in \mathbb{R}^{N_B \times dim}$. The loss functions are defined as follows:

\subsubsection{Detector Loss} utilizes the reprojection loss. Since the extracted keypoints are differentiable, their positions can be directly optimized using the reprojection distance through the forward and backward propagation. For a keypoint \textit{\textbf{p}}$_A$ from image \textit{\textbf{I}}$_A$, its position is wrapped into image \textit{\textbf{I}}$_B$ by 3D perspective projection:
\begin{equation}
    p_{AB} = \mathrm{\mathbf{wrap}}_{AB}(p_A) = \pi(d_A R_{AB} \pi^{-1}(p_A) + t_{AB}),
\end{equation}
where the rotation matrix \textbf{\textit{R}}$_{AB}$ and translation vector \textbf{\textit{t}}$_{AB}$ describe the transformation from \textbf{\textit{I}}$_A$ to \textbf{\textit{I}}$_B$. The term \textbf{\textit{d}}$_A$ denotes the depth of keypoint \textit{\textbf{p}}$_A$, and $\boldsymbol \pi(\cdot)$ represents the operation that projects a three-dimensional point onto the image plane. In the image \textbf{\textit{I}}$_B$, the nearest keypoint to the projected position \textbf{\textit{p}}$_{AB}$ is identified as \textbf{\textit{p}}$_B$, provided their distance is within $th_{gt}$ pixels. The \textbf{\textit{p}}$_B$ is regarded as the corresponding match for \textit{\textbf{p}}$_A$. Likewise, \textbf{\textit{p}}$_B$ is reprojected back into \textbf{\textit{I}}$_A$, resulting in \textbf{\textit{p}}$_{BA}$. The detector loss of the keypoint pair is defined as:
\begin{equation}
    \mathcal{L}_{\mathrm{det}}(p_A, p_B) = \frac{1}{2} \left( \| p_A - p_{BA} \| + \| p_B - p_{AB} \| \right).
\end{equation}

The detector loss, denoted as $\mathcal{L}_{det}$, is computed by averaging the reprojection loss across all corresponding keypoints between the two images.

\subsubsection{Descriptor Loss} abandons the classical triplet loss and adopts the neural reprojection error (NRE) loss~\cite{germain2021neural} to optimize the descriptor map. NRE evaluates the discrepancy between two probability maps named matching probability map and reprojection probability map. This imposes a strong constraint on the dense descriptor map, ensuring a reliable training process.

For the given keypoint \textbf{\textit{p}}$_A$ within \textbf{\textit{I}}$_A$ and its corresponding keypoint \textbf{\textit{p}}$_{AB}$ within \textbf{\textit{I}}$_B$, their reprojection probability map is established through the application of bilinear interpolation, as delineated in reference~\cite{zhao2022alike}, resulting in the expression: \textbf{\textit{q}}$_r$$($\textbf{\textit{p}}$_B|$\textbf{\textit{p}}$_{AB})$. For the descriptor \textbf{\textit{d}}$_{p_A}$ with the descriptor map \textbf{\textit{D}}$_B$, their matching probability map is given as the softmax normalization:
\begin{equation}
    q_m(p_B \mid d_{p_A}, D_B) := \mathrm{softmax} \left( \frac{D_B \cdot d_{p_A} - 1}{t_{des}} \right),
\end{equation}
where $t_{des}$ denotes the sharpness of the matching probability map, which influences the the concentration of the probability distribution around the reprojected keypoint \textbf{\textit{p}}$_{AB}$. NRE optimizes the discrepancy across the dense reprojection and the matching probability maps using cross-entropy (CE) loss function:
\begin{align}
    NRE(p_A, I_B) 
    &:= \mathrm{CE} \left( q_r(p_B \mid p_{AB}) \parallel q_m(p_B \mid d_{p_A}, D_B) \right) \notag \\
    &= - \ln \left( q_m(p_{AB} \mid d_{p_A}, D_B) \right).
\end{align}

Thus, the descriptor loss can be defined in a symmetric way:
\begin{equation}
    \mathcal{L}_{\mathrm{des}} = \frac{1}{N_A + N_B} \left( 
    \sum_{p_A \in I_A} \mathrm{NRE}(p_A, I_B) + \sum_{p_B \in I_B} \mathrm{NRE}(p_B, I_A) 
    \right),
\end{equation}
where $N_x$ represents the counts of detected keypoints within the input image.

\subsubsection{Coupling Loss} constraints the keypoints in discriminative areas to improve their reliability with the assistance of descriptors. Inspired by methods following the `detect-and-describe' paradigm like D2-Net~\cite{dusmanu2019d2}, we define it as:
\begin{equation}
    \mathcal{L}_{\mathrm{cp}}^A = \frac{1}{N_A} \sum_{\substack{p_A \in I_A \\ p_{AB} \in I_B}} \frac{s_{p_A} s_{p_{AB}}}{\sum_{\substack{p'_A \in I_A \\ p'_{AB} \in I_B}} s_{p'_A} s_{p'_{AB}}} \mathcal{M}(p_{AB}, d_{p_A}, D_B),
\end{equation}
where $N_A$ represents the counts of detected keypoints within image \textbf{\textit{I}}$_A$. And \textbf{\textit{s}}$_p$ represents the score of the keypoint \textbf{\textit{p}}. Specially, we define the ranking loss $\mathcal{M}(\hspace{0.3pt} \cdot \hspace{0.3pt})$ for representation learning as:
\begin{equation}
    \mathcal{M}(p_{AB}, d_{p_A}, D_B) = 1 - \mathcal{S} \left( \exp \left( \frac{D_B \cdot d_{p_A} - 1}{t_{des}} \right), p_{AB} \right),
\end{equation}
where $\mathcal{S}(\cdot, \cdot)$ is the bilinear sampling at position $\in \mathbb{R}^2$ and probability map $\in \mathbb{R}^{H \times W}$. Similarly, the overall coupling loss is given symmetrically as:
\begin{equation}
    \mathcal{L}_{\mathrm{cp}} = \frac{1}{2} \left( \mathcal{L}_{\mathrm{cp}}^A + \mathcal{L}_{\mathrm{cp}}^B \right).
\end{equation}

\begin{figure}[t]
	\centering
	\includegraphics[width=0.97\linewidth]{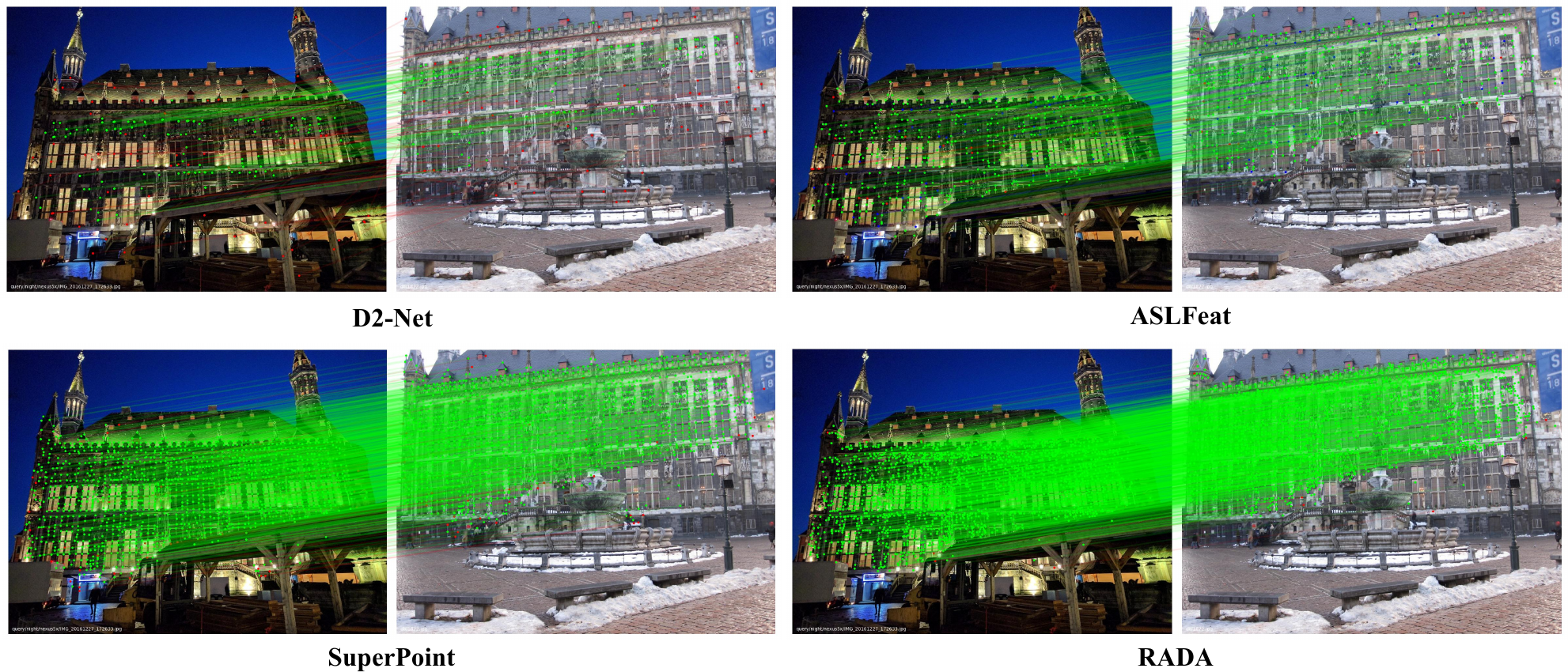}
	\caption{
	  Visualization of matches on Aachen Day-Night~\cite{zhang2021reference}. The color-coded inliers: green for correct matches (reprojection error within 0 to 5 pixels), red for incorrect matches (exceeding 5 pixels), blue for unavailable ground truth depth. }
	\label{fig:results}
\end{figure}

\subsubsection{Total Loss} combines these loss functions, each with its own coefficient:
\begin{equation}
\label{eq:overall_loss}
    \mathcal{L}_{\mathrm{total}} = \omega_{\mathrm{da}} \mathcal{L}_{\mathrm{da}} + \omega_{\mathrm{Tr}} \mathcal{L}_{\mathrm{Tr}} + \omega_{\mathrm{det}} \mathcal{L}_{\mathrm{det}} + \omega_{\mathrm{des}} \mathcal{L}_{\mathrm{des}} + \omega_{\mathrm{cp}} \mathcal{L}_{\mathrm{cp}}.
\end{equation}

%
%
\section{Experiments}

\subsection{Implementation Details}
\subsubsection{Training Data:} We trained RADA on the MegaDepth dataset~\cite{li2018megadepth} and adopted the scenes employed for training in DISK~\cite{tyszkiewicz2020disk}. Following the cross-domain data preparation strategy proposed in~\cite{xu2023domainfeat},we used HiDT~\cite{anokhin2020high} to translate the source domain images \textbf{\textit{I}}$_S$ into dusk or evening domains images \textbf{\textit{I}}$_T$. Since the target domain images are also sourced from MegaDepth, their corresponding ground truth can be retrieved. Following the method outlined in D2-Net~\cite{dusmanu2019d2}, we computed the overlap score between pairs of images. For each scene in every epoch, we then selected 100 training pairs with overlap scores within the range of $[0.3, 1]$. 

\subsubsection{Training Details:} The radius of the DKD module was set to 2 pixels, and keypoints were considered ground truth pairs if their reprojection distances were less than 5 pixels. The loss coefficients in~\cref{eq:overall_loss}, $(\boldsymbol\omega_{da},\space \boldsymbol\omega_{Tr},\space \boldsymbol\omega_{det},\space \boldsymbol\omega_{des},\space \boldsymbol\omega_{cp})$ were optimized by Optuna~\cite{akiba2019optuna} as $(2,1,1,5,1)$, and $\lambda$ in~\cref{eq:da} was set to 0.01. The input images were initially cropped and scaled to $480 \times 480$ pixels. During training, the ADAM optimizer was employed to adjust the learning rate from $0$ to $3e^{-3}$ over 500 steps before stabilizing. We set the batch size as 2, with gradient accumulation over 16 batches. Under these conditions, the model converged in approximately 30 hours on a single NVIDIA RTX 4090.

\subsection{Image Matching}

\begin{table}[b]
  \caption{Image matching performance on HPatches~\cite{balntas2017hpatches}. GFLOPs and FPS for 640 × 480 images, along with illumination and overall MMA within the threshold range of one to three pixels for various methods. The best three results are highlighted as \textcolor{darkred}{\uuline{red}}, \textcolor{darkgreen}{\uline{green}}, \textcolor{darkblue}{\uwave{blue}}. \textit{*} represents similar results.}
  \label{tab:hpatches}
  \centering
  \scriptsize
  \begin{tabular}{l@{\hspace{0.02cm}}|c@{\hspace{0.02cm}}c@{\hspace{0.1cm}}|c@{\hspace{0.05cm}}c@{\hspace{0.1cm}}c@{\hspace{0.03cm}}|@{\hspace{0.08cm}}c@{\hspace{0.18cm}}c@{\hspace{0.18cm}}c@{}}
    \toprule
    \textbf{Models}&\textit{GFLOPs}&\textit{FPS}&\textit{MMA\_i@1}&\textit{MMA\_i@2}& \textit{MMA\_i@3}&\textit{MMA@1}&\textit{MMA@2}& \textit{MMA@3}\\
    \midrule
    {\bfseries D2-Net}~\cite{dusmanu2019d2}&889.40 &6.60 &13.76\%& 29.01\%& 43.22\%&9.78\%& 23.52\%& 37.29\%\\
    {\bfseries SuperPoint}~\cite{detone2018superpoint}&\textcolor{darkred}{\uuline{26.11}}&\textcolor{darkred}{\uuline{45.87}}&43.09\%& 59.39\%& 69.25\%&34.27\%& 54.94\%& 66.37\%\\
    {\bfseries R2D2}~\cite{revaud2019r2d2}&464.55 &8.70 &37.60\%& 66.08\%& \textcolor{darkgreen}{\uline{80.78\%}}&33.31\%& 62.17\%& \textcolor{darkblue}{\uwave{75.77\%}}\\
 {\bfseries ASLFeat}~\cite{luo2020aslfeat}&\textcolor{darkgreen}{\uline{44.24}}&8.96 &46.12\%& 64.46\%& 75.45\%& 39.16\%&  61.07\%& 72.44\%\\
 {\bfseries DISK}~\cite{tyszkiewicz2020disk}&98.97 &15.73 & \textcolor{darkgreen}{\uline{51.98\%}}&\textcolor{darkred}{\uuline{77.11\%}}& \textcolor{darkred}{\uuline{82.61\%}}& \textcolor{darkblue}{\uwave{43.71\%}}&\textcolor{darkgreen}{\uline{66.98\%}}& \textcolor{darkred}{\uuline{77.59\%}}\\
 {\bfseries ALIKE}~\cite{zhao2022alike}&\textcolor{darkblue}{\uwave{67.63*}}&\textcolor{darkgreen}{\uline{29.15}}&\textcolor{darkblue}{\uwave{51.24\%}}& \textcolor{darkblue}{\uwave{69.51\%}}& 77.40\%&\textcolor{darkgreen}{\uline{44.97\%}}& \textcolor{darkblue}{\uwave{66.21\%}}& 74.51\%\\
   {\bfseries RADA(ours)}&\textcolor{darkblue}{\uwave{67.74*}}&\textcolor{darkblue}{\uwave{24.56}}&\textcolor{darkred}{\uuline{52.13\%}}& \textcolor{darkgreen}{\uline{70.83\%}}& \textcolor{darkblue}{\uwave{78.65\%}}& \textcolor{darkred}{\uuline{45.03\%}}&\textcolor{darkred}{\uuline{67.17\%}}& \textcolor{darkgreen}{\uline{76.24\%}}\\
    \midrule
  \end{tabular}
\end{table}

\subsubsection{Experiment setup:} HPatches~\cite{balntas2017hpatches} is a widely used benchmark for the image matching task, which contains 116 scenes and 580 image pairs with known ground truth homography. Following the protocol of D2-Net, we exclude 8 high-resolution sequences from the origin and employ 108 scenes with substantial variations in illumination or viewpoints. We record mean matching accuracy (MMA), the average ratio of accurate matches to all estimated putative matches, under thresholds ranging from one to ten pixels, along with the quantities of features and matches. We also report GFLOPs and the interface FPS to represent the computing complexity and the response speed. As D2-Net, mutual nearest neighbor search is employed during the latter stages of the matching process.

\subsubsection{Results:} Table~\ref{tab:hpatches} shows the results on HPatches under challenging domain changes, such as illumination and viewpoints. Our method generally outperforms other proposed methods. Although R2D2~\cite{revaud2019r2d2} and DISK~\cite{tyszkiewicz2020disk} sometimes perform better on different evaluation metrics, they sacrifice either the computing complexity or the response speed. RADA balances accuracy, computing complexity, and response speed organically, resulting in better overall performance. 

\subsubsection{Ablation:} We apply ablation studies on HPatches to explore the performance of our special network designs and architecture loss functions. We introduce two additional evaluation metrics: 
matching score (M.S.), which measures the percentage of correct matches among keypoints in shared views, and mean homography accuracy (MHA), defined as the average ratio of accurately identified image corners to the estimated homography matrix. The results are presented in Table~\ref{tab:ablation}.
The domain adaptation supervision ensures the network learns domain-invariant features, and the Transformer-based booster enhances the accuracy and robustness of features, consistent with the ablation study outcomes. The detector loss optimizes the positions of keypoints to enhance their accuracy, while the coupling loss guarantees that the detected keypoints are positioned in regions where existing highly discriminative descriptors. Moreover, the Wave-PE enhances the network's accuracy and the AFT significantly decreases the computational complexity. These designs lead to more trustworthy keypoints, which reduces the occurrence of false matches and improves overall matching accuracy.

\begin{table*}[t]
\caption{Ablation studies on network architecture and losses. \textit{DET}, \textit{DES}, and \textit{CP} denote detector loss, descriptor loss and coupling loss, respectively.}
\label{tab:ablation}
\centering
\caption*{(a) Ablation study on main network architecture designs.}
\label{tab: architecture ablation}
\resizebox{.88\textwidth}{!}{
 \begin{tabular}{@{\hspace{0.05cm}}l@{\hspace{0.85cm}}c@{\hspace{0.8cm}}c@{\hspace{0.8cm}}c@{\hspace{0.05cm}}}
		\toprule
		\textbf{Methods  }    	& M.S.    & MMA@3    & MHA@3    \\ \midrule
		\textbf{Backbone  }    & 36.96\%      & 70.48\% & 68.26\% \\
		\textbf{ + Domain Adaptation Supervision} & 38.09\%  & 74.72\% & 68.44\%\\
		\textbf{ + Transformer-based Booster} 	  & \textbf{38.74\%}  & \textbf{76.24\%} & \textbf{70.46\%} \\ \bottomrule
	\end{tabular}%
}
\vspace{0.5cm}
\begin{minipage}{0.4\linewidth}
  \centering
  \caption*{(b) Ablation study on losses.}
  \label{tab: loss ablation}
  \resizebox{.92\columnwidth}{!}{
 \begin{tabular}{@{}c@{\hspace{0.2cm}}c@{\hspace{0.2cm}}c@{\hspace{0.2cm}}c@{\hspace{0.25cm}}c@{}}
	\toprule
	\textbf{DET} & \textbf{DES} &\textbf{CP} & MMA@3 & MHA@3 \\ \midrule
	&    \checkmark& \checkmark              & 71.86\%  & 67.84\%\\
	 \checkmark  & \checkmark      &       & 68.92\%   & 66.98\%\\
	 \checkmark  &        \checkmark& \checkmark        & \textbf{76.24\%} & \textbf{70.46\%} \\ \bottomrule
	\end{tabular}%
  }
\end{minipage}
\hfill
\begin{minipage}{0.58\linewidth}
  \centering
  \caption*{(c) Ablation study on booster designs.}
  \label{tab: attention ablation}
  \resizebox{.94\columnwidth}{!}{
   \begin{tabular}{@{}c@{\hspace{0.2cm}}c@{\hspace{0.2cm}}c@{\hspace{0.2cm}}c@{\hspace{0.2cm}}c@{\hspace{0.2cm}}c@{}}
	\toprule
	MLP-PE & \textbf{Wave-PE} &MHA & \textbf{AFT} & GFLOPs & MMA@3\\ \midrule
	 \checkmark &  & \checkmark &  & 101.46 & 78.15\%\\
	 & \checkmark & \checkmark &   & 109.63 & \textbf{78.92\%}\\
	 & \checkmark &  & \checkmark & \textbf{67.74}  & 76.24\%\\ \bottomrule
  \end{tabular}%
  }
\end{minipage}
\end{table*}

\subsection{Visual Localization}

\subsubsection{Experiment setup:}
We adopt Aachen Day-Night v1.1~\cite{zhang2021reference}, an outdoor dataset characterized by significant illumination changes, containing 6,697 daytime data-base images along with 1,015 query images (824 daytime and 191 nighttime) for visual localization. We employ the hierarchical localization toolbox (HLoc)\cite{sarlin2019coarse} and Long-Term Visual Localization Benchmark\cite{toft2020long} to record the percentage of query images correctly localized within specified error thresholds. The whole methods adopt mutual nearest neighbor search for the matching task, and we apply the distance and ratio test for this, with thresholds consistent with~\cite{wang2023featurebooster}.

\subsubsection{Results:}
The results are shown in Table~\ref{tab:aachen}. This demonstrates that our descriptor, enhanced through domain adaptation, significantly improves performance and effectively integrates with keypoint detection, even under challenging day-night domain changes. As shown in Fig.~\ref{fig:results}, our method can detect more accurate keypoints and generate effective matches under the same conditions, and the descriptors of keypoint locations have higher discriminability. Thus, our learning descriptors with domain adaptation can substantially improve the performance of visual localization tasks that are fraught with domain-shifted challenges.

\begin{table}[h]
\caption{Visual localization performance on Aachen Day-Night v1.1~\cite{zhang2021reference}. We report the positional and angular performance metrics, with higher values indicating better performance. The top two results are highlighted with \textcolor{darkred}{\uuline{red}}, \textcolor{darkgreen}{\uline{green}}.
  }
\label{tab:aachen}
\centering
\resizebox{0.7\textwidth}{!}{
\scriptsize
\begin{tabular}{@{\hspace{0.05cm}}l@{\hspace{0.5cm}}|@{\hspace{0.65cm}}c@{\hspace{1.3 cm}}c@{\hspace{0.15cm}}}
\toprule
\multirow{2}{*}{\textbf{Method}} 
& Day & Night \\
    & \multicolumn{2}{c}{\textit{(0.25m,2$^{\circ}$) / (0.50m,5$^{\circ}$) / (5.0m,10$^{\circ}$)}} \\ 
\midrule    
{\bfseries SIFT}\cite{lowe2004distinctive} & 87.1 / 93.8 / 98.1 & 50.8 / 70.2 / 81.2 \\
{\bfseries SuperPoint}\cite{detone2018superpoint} & 87.9 / 94.3 / 98.2& 67.0 / 84.8 / 95.8\\
{\bfseries SOSNet}\cite{tian2019sosnet} & \textcolor{darkred}{\uuline{88.7}} / \textcolor{darkred}{\uuline{94.7}} / 98.7& 58.1 / 78.5 / 92.7 \\
\textbf{R2D2}\cite{revaud2019r2d2} & 86.8 / 94.1 / 98.4& 57.1 / 76.4 / 88.5 \\
{\bfseries DISK}\cite{tyszkiewicz2020disk} & 86.9 / 93.3 / \textcolor{darkred}{\uuline{99.2}}&\textcolor{darkgreen}{\uuline{69.4}} / \textcolor{darkgreen}{\uuline{86.8}} / 96.7\\
{\bfseries ALIKE}\cite{zhao2022alike} & 87.3 / 93.2 / 98.7& 67.5 / 85.3 / \textcolor{darkgreen}{\uuline{97.9}}\\
{\bfseries RADA(ours)} & \textcolor{darkgreen}{\uuline{87.9}} / \textcolor{darkgreen}{\uuline{94.4}} / \textcolor{darkgreen}{\uuline{99.0}}& \textcolor{darkred}{\uuline{72.9}} / \textcolor{darkred}{\uuline{86.8}} / \textcolor{darkred}{\uuline{98.6}} \\
\bottomrule
\end{tabular}}
\end{table}

\section{Conclusion}
In this study, we develop RADA, a multi-level feature aggregation network incorporating domain adaptation supervision and Transformer-based booster, to learn robust and accurate features under extreme domain shifts. The hierarchical network efficiently captures diverse features to achieve accurate keypoint detection and descriptor extraction. By aligning high-level feature distributions across different domains, the domain adaptation supervision ensures the learning of invariant domain representations. The Transformer-based booster integrating visual and geometric information is proposed to enhance the robustness of descriptors. To ensure the accuracy and robustness of features, we propose three additional losses to apply targeted supervision on keypoint detection, descriptor extraction, and their coupled processing. Extensive experiments conducted in image matching and visual localization tasks validate the superiority and effectiveness of our method among these applications.

\bibliographystyle{splncs04}
\bibliography{1672}

\end{document}